\pdfoutput=1

\documentclass[11pt]{article}

\usepackage{ACL2023}

\usepackage{times}
\usepackage{latexsym}

\usepackage[T1]{fontenc}

\usepackage[utf8]{inputenc}

\usepackage{microtype}

\usepackage{inconsolata}

\usepackage{microtype}
\usepackage{booktabs}       
\usepackage{amsmath}
\usepackage{amsfonts}
\usepackage{array}
\usepackage{bm}
\usepackage{multirow}
\usepackage{xcolor}         
\usepackage{subcaption}
\usepackage{hyperref}       
\usepackage{url}            
\usepackage{graphicx} 
\title{Learning Better Masking for Better Language Model Pre-training}

\author{Dongjie Yang\textsuperscript{\rm 1,2},
    Zhuosheng Zhang\textsuperscript{\rm 1,2,}\thanks{\;\;Corresponding author; This paper was partially supported by Key Projects of National Natural Science Foundation of China (U1836222 and 61733011).
    }\,\,,
    Hai Zhao\textsuperscript{\rm 1,2,}\footnotemark[1] \\
    \textsuperscript{\rm 1}Department of Computer Science and Engineering, Shanghai Jiao Tong University\\
    \textsuperscript{\rm 2}Key Laboratory of Shanghai Education Commission for Intelligent Interaction \\ and Cognitive Engineering, Shanghai Jiao Tong University\\
    \texttt{\{djyang.tony,zhangzs\}@sjtu.edu.cn,zhaohai@cs.sjtu.edu.cn}}

\begin{document}
\maketitle
\begin{abstract}
Masked Language Modeling (MLM) has been widely used as the denoising objective in pre-training language models (PrLMs). Existing PrLMs commonly adopt a Random-Token Masking strategy where a fixed masking ratio is applied and different contents are masked by an equal probability throughout the entire training. However, the model may receive a complicated impact from pre-training status, which changes accordingly as training time goes on. In this paper, we show that such time-invariant MLM settings on masking ratio and masked content are unlikely to deliver an optimal outcome, which motivates us to explore the influence of time-variant MLM settings. We propose two scheduled masking approaches that adaptively tune the masking ratio and masked content in different training stages, which improves the pre-training efficiency and effectiveness verified on the downstream tasks. Our work is a pioneer study on time-variant masking strategy on ratio and content and gives a better understanding of how masking ratio and masked content influence the MLM pre-training\footnote{\url{https://github.com/mutonix/better_masking}}.
\end{abstract}

\section{Introduction}
Pre-trained language models (PrLMs) have played an essential role in many natural language processing tasks \citep{radford2018improving,devlin-etal-2019-bert,bao2020unilmv2,guu2020realm,yu2021dict,zhang2022opt}. Generally speaking, PrLMs can be seen as an automatic denoising encoder and may be conveniently obtained through a self-supervised learning way. Masked Language Modeling (MLM) pioneered by BERT \citep{devlin-etal-2019-bert} is a widely used denoising method for language model pre-training \citep{lan2019albert,clark2019electra}. In MLM pre-training, a subset of tokens in a sequence is masked with a certain masking ratio, and the masked sequence is fed to the PrLM, which is required to predict the masked tokens.

Masking in MLM is a process in terms of sampling masked tokens from a huge data space to generate training batches, in which MLM may be heavily controlled by two main factors, masking ratio and masked contents. So far, only a few studies have ever considered optimal settings for better MLM from quite limited perspectives. Especially, all known works only take time-invariant MLM setting into account despite the huge time variance of the model during a lengthy pre-training. For example, carefully considered masked units like $n$-gram, entity and span \citep{sun2019ernie,joshi-etal-2020-spanbert,levine2021pmimasking,li-zhao-2021-pre} are adopted throughout the entire pre-training. Another example is that exploring a good enough (but still fixed) masking ratio has also been considered in \citep{wettig2022should}. Given the circumstances that time-invariant masking applied in most MLM is not adaptive to the changeable process of language model pre-training, time-invariant setting hardly hopefully reaches an optimal outcome. This motivates us to explore the influence of masking ratio and masked content in MLM pre-training and propose time-variance MLM setting to verify our hypothesis for better PrLMs.

\begin{table*}[ht]
 \setlength{\tabcolsep}{4pt}
 \centering
\caption{Examples of masking function words and non-function words. Intuitively, it is much easier to predict the masked function words.}
\label{tab:mis_example}
\vskip 0.1in \small
\begin{tabular}{lllcc}
\toprule
\textbf{\textbf{Masking Strategy}} & \textbf{Example} \\
\midrule
  Function words &  
  \colorbox[RGB]{221,226,237}{{[MASK]}} \quad apple \quad  \colorbox[RGB]{221,226,237}{{[MASK]}}\quad day \quad keeps \quad \colorbox[RGB]{221,226,237}{{[MASK]}} \quad doctor \quad \colorbox[RGB]{221,226,237}{{[MASK]}} . \\
  Non-function words & \quad An \qquad \colorbox[RGB]{253,232,216}{{[MASK]}} \:\quad a \:\quad \colorbox[RGB]{253,232,216}{{[MASK]}} \colorbox[RGB]{253,232,216}{{[MASK]}} \quad the \quad \colorbox[RGB]{253,232,216}{{[MASK]}} \quad away.   \\
\bottomrule
\end{tabular}
\end{table*}

\textbf{$\bullet$ Masking Ratio}. 
Masking ratio controls the ratio between the number of tokens to predict and the left corrupted context. It determines the corruption degree that may affect the difficulty of restoring the masked tokens; that is, the larger the ratio is, the more masked contents model has to predict with less non-masked context. Our hypothesis is that at different training stages, the model may benefit from different masking ratios to balance the training from samples with different difficulties compared to the fixed ratio.

We first explore the influence of different masking ratios on downstream tasks at different stages throughout the entire pre-training instead of the only final stage. We find that a high masking ratio gives better performance for downstream tasks in early stage, while a low ratio has a faster training speed. Thus we choose a higher ratio as the starting point and decay the masking ratio to a lower value during the pre-training, namely Masking Ratio Decay (MRD), which can significantly outperform the performance of the fixed ratio. MRD indicates that MLM benefits from a time-variant masking ratio at different training stages.

\textbf{$\bullet$ Masked Content}. When placing all words with an equal and fixed opportunity throughout the entire pre-training for prediction learning, it may be unnecessary for some 'easy' words and insufficient for some 'difficult' words at the same time. Table \ref{tab:mis_example} shows an intuitive example that the sequence with masked non-function words containing less information is much harder to predict compared to masked function words. Though in the very beginning, all words are unfamiliar to the models. As time goes on, the relative difficulties of words will vary when the pre-training status changes. We show that the losses of function words converge much faster than non-function words, which means non-function words are much harder for models to learn. Therefore, the high proportion of function words in a sequence leads to inefficiency if Random-Token Masking is applied.

To handle training maturity for different types of words, we propose POS-Tagging Weighted (PTW) Masking to adaptively adjust the masking probabilities of different types of words according to the current training state. PTW Masking makes the model have more chance to learn 'difficult' types of words and less chance for 'easy' ones from the perspective of part-of-speech. By introducing this adaptive schedule, our experimental results show that MLM benefits from learning mostly non-function words.



Our contributions are three folds:
1) We analyze the insufficiency of current masking strategies from the perspectives of masking ratio and masked content and give a better understanding of MLM pre-training in terms of masking.
2) To our best knowledge, this is a pioneer study to analyze the impact of time-variant masking both in masking ratio and masked content in MLM pre-training.
3) Our analysis shows that the time-variant masking schedules can significantly improve training efficiency and effectiveness.
Our sources will be publicly available.

\section{Preliminary Experiments}
This section presents our preliminary experiments that motivate us to explore time-variant masking schedules. We train BERT-base \citep{devlin-etal-2019-bert} with the widely-used English Wikipedia corpus to observe the influence of masking during pre-training by measuring the downstream performance on the SQuAD v1.1 dataset \citep{Rajpurkar2016SQuAD} (more experimental details will be given in Section \ref{sec:exp}). The experiments aim to study how the language model learns from the masked tokens when using conventional Random-Token Masking from the perspectives of the masking ratio and masked content.

\subsection{Preliminaries of Masked Language Model}
In general, Masked Language Modeling (MLM) is a denoising auto-encoding approach that is widely used in language model pre-training by reconstructing the corrupted sequences. To be specific, given a sequence $\bm{x}=\{x_{i},x_{2},\dots ,x_{n}\}$, we use a certain masking strategy $\bm{P}$ to replace $p\%$ tokens with special mask tokens. Accepting the corrupted sequence as input, a language model parameterized by $\theta$ is trained to predict the original tokens from masked ones in $\bm{x}$ using the pre-training objective stated below:
\begin{equation}\label{eq:mlm_loss}
    \mathcal{L}_{\mathrm{MLM}}(\bm{x}, \theta) = \mathbb{E}\Biggl(-\sum_{i\in \mathbb{M}} log\: p_{\theta}(x_{i}\mid \hat{\bm{x}})\Biggr),
\end{equation}
where $\hat{\bm{x}}$ is the reconstructed sequence that the language model samples from the hidden states, and $\mathbb{M}$ denotes the index set of masked tokens where the loss will be calculated.


\begin{figure*}
    \centering
    \begin{subfigure}{0.3\textwidth}
	\centering
	\includegraphics[width=1\textwidth]{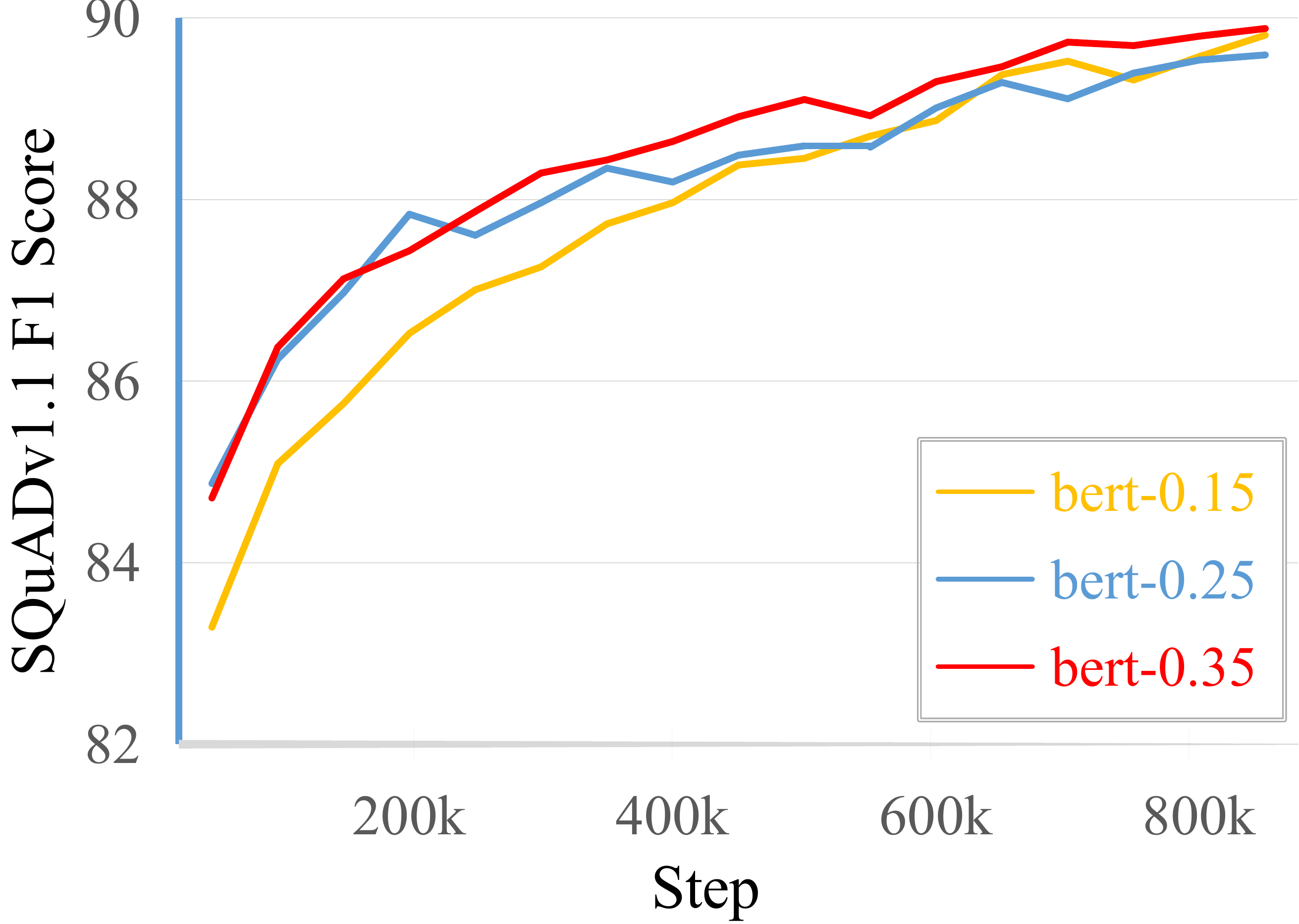}
	\caption{SQuAD v1.1 performance using different masking ratios.}
	\label{fig:mlm_ratio}
    \end{subfigure}
    \quad
    \begin{subfigure}{0.3\textwidth}
	\centering
	\includegraphics[width=1\textwidth]{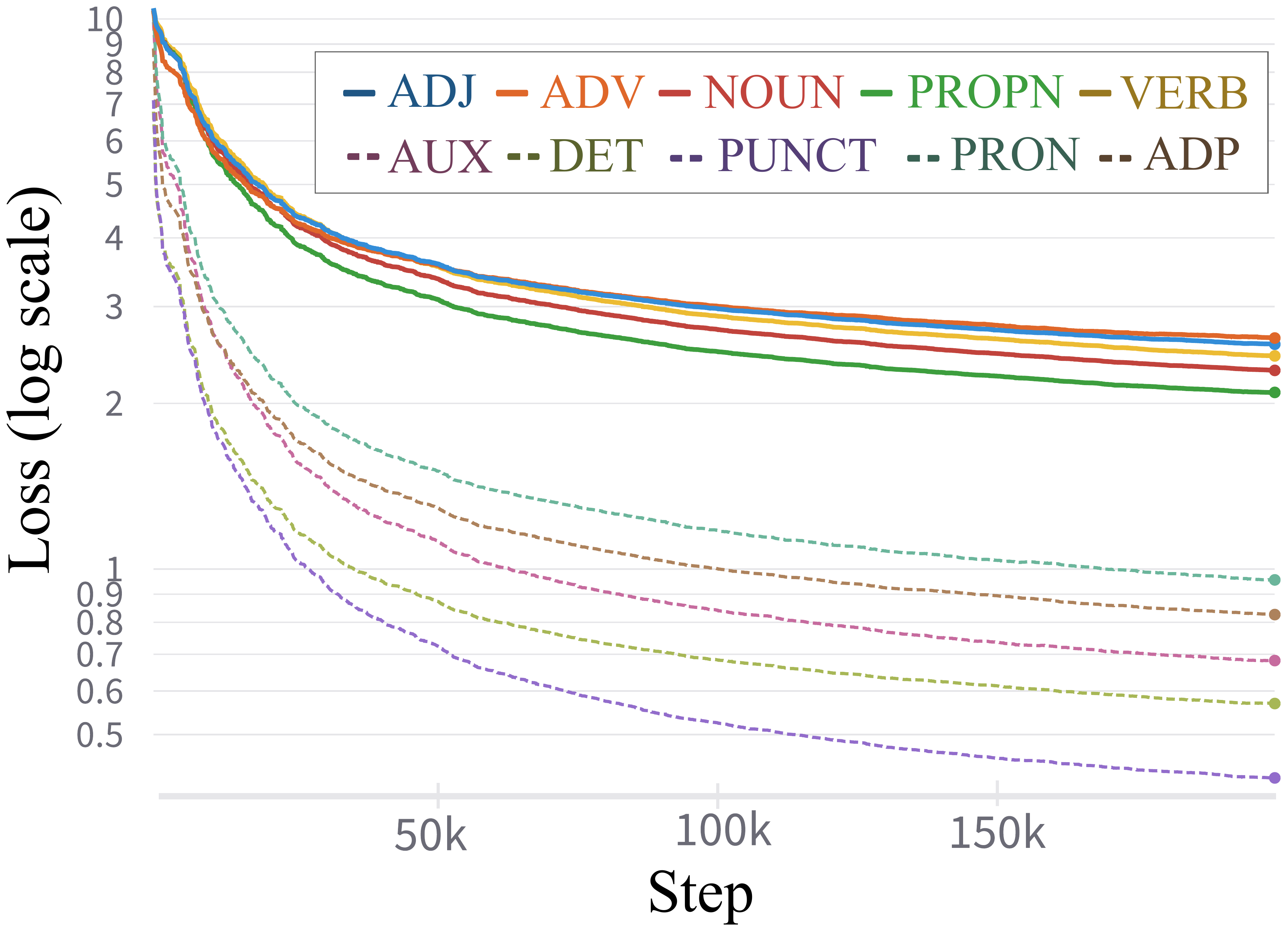}
	\caption{Cumulative losses of different types of words (shown 10 of them).}
	\label{fig:pos}
    \end{subfigure}
    \quad
    \begin{subfigure}{0.3\textwidth}
	\centering
	\includegraphics[width=1\textwidth]{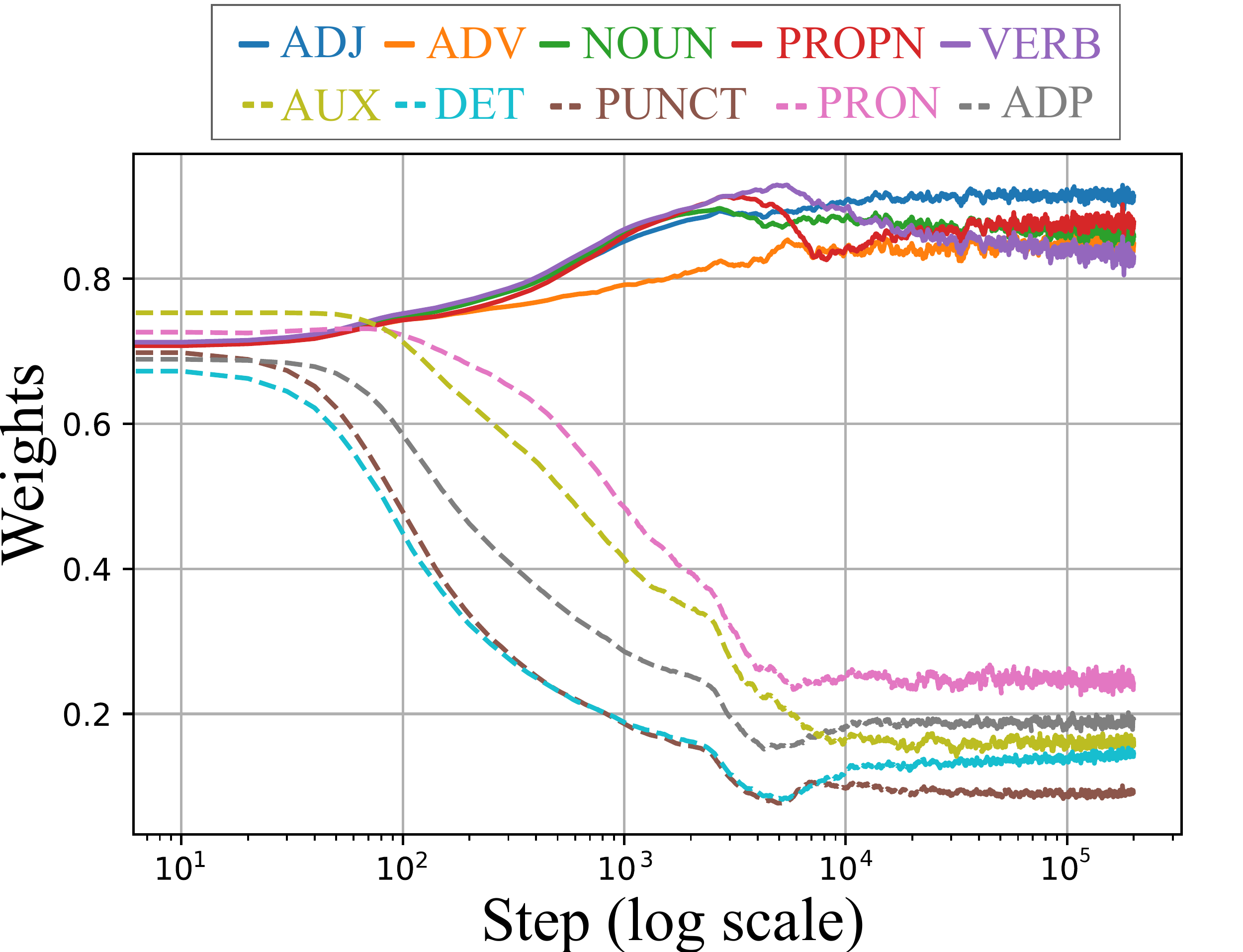}
	\caption{Weighting changes of different types of words (shown 10 of them).}
	\label{fig:weights}
    \end{subfigure}
    \caption{(a) Experiment on masking ratio: The evaluation metric F1 score measures the average textual overlap between the prediction and ground truth answer. (b) Experiment on masked content: Cumulative losses of different types of words (shown 10 of them). The losses of non-function words (in full line) are much greater than the losses of function words and punctuation (in dotted line). (c) Weight changes of PTW according to the cumulative losses in (b): Based on Equation \ref{eq:pos_wt}, the masking weights of non-function words (in full line) are much higher than the weights of function words and punctuation (in dotted line).}
\end{figure*}

\subsection{Masking Ratio: Influence of Pre-training Masking Ratio}
\label{sec:mlm_ratio}

The masking ratio determines the corruption degree of a whole sequence for model training. We first conduct a simple experiment to explore the impact of different masking ratios on downstream tasks at different training stages of entire pre-training.


We train BERT-base for 1M steps with a masking ratio of 15\%, 25\%, and 35\%  respectively, as shown in Figure \ref{fig:mlm_ratio}. During pre-training, checkpoints are saved every 50k steps, and finetuning is performed on the SQuAD v1.1 to observe changes in downstream performance. We find that the models using masking ratios of 25\% and 35\%  have a gap of more than +1\% F1 score in SQuAD compared with the model using a masking ratio of 15\%  at the beginning of training. However, in the second half training stage, the model with the masking ratio of 15\% catches up with models with higher ratios in downstream performance.

\subsection{Masked Content: Influence of Different Types of Words} 
\label{sec:pos}

In this section, we observe the influence of masked content by finding which kinds of words are more beneficial to pre-training. In terms of part-of-speech, words can be roughly divided into three categories: non-function words, function words, and the others (punctuations, symbols, etc.).
If we mask all the function words and punctuations of a sentence, we can still infer roughly what the sentence is about. Instead, by masking all non-function words, we can hardly get any information from the sentence, as shown in Table \ref{tab:mis_example}. 

To further explore the part-of-speech, we can speculate that, for the language model, masking different types of words leads to different difficulties for pre-training. 

With the help of POS-tagging tools\footnote{POS-tagging tool in spaCy: \url{https://spacy.io/}}, we classify the words in the corpus into $m$ categories\footnote{Part-of-speech classification from the Universal POS tag set: \url{https://universaldependencies.org/u/pos/}} when doing pre-processing. In pre-training, for each type of words, we calculate the corresponding cumulative loss $\tilde{\ell}_{k,\:t}$ at $t$ steps as follows:
\begin{equation}
\label{eq:pos}
\begin{gathered}
    \mathcal{L}_{\mathrm{MLM}}(\bm{x}, \theta) = \sum_{k\in \mathcal{C}}\ell_{k,\:t}\:, \\
    \tilde{\ell}_{k,\:t} = \beta \cdot \tilde{\ell}_{k,\:t-1} + (1-\beta) \cdot \ell_{k,\:t}\:,
\end{gathered}
\end{equation}
where $k \in \mathcal{C}$ denotes the word type $k$ in set $\mathcal{C}$ of $m$ categories and $\beta \in (0,1)$ is a coefficient to balance the exponential weighted average. We use exponential weighted average to smooth the losses because temporary losses of different batches vary greatly, leading to corresponding weights jittering (more details will be discussed in Section \ref{sec:ptw_masking}).


We train BERT-base for 200k steps with a fixed masking ratio of 15\%. We record the cumulative losses of different types of words separately every 10 steps and observe the changes in losses. We find that the language model does have higher losses for masked non-function words and lower losses for masked function words. The latter quickly converges to very small values from the start, as shown in Figure \ref{fig:pos}.

\subsection{Analysis}
\label{sec:analysis}
\paragraph{Masking Ratio: Why Time-invariant Masking Ratio Is Not the Best Choice?}
From the experimental results in Figure \ref{fig:mlm_ratio}, there is such an empirical law: at the beginning, the downstream performance with a high masking ratio has a higher starting point but grows at a relatively slower speed and is caught up with the model with masking ratio of 15\%. That is, the model with the masking ratio of 15\%  has a low starting point but boosts performance faster in the later stage. Given this observation, we show that we can apply a relatively high masking ratio to train models to get a better model using less time. On the other hand, we apply a lower masking ratio to train models, which obtains better downstream performance if we train for enough time. But if we use a decaying masking ratio instead of a fixed one, we can absorb the advantages of both high and low masking ratios.

\paragraph{Masked Content: Why Random-Token Masking Is Suboptimal?}
For a sentence, the numbers of non-function words and function words are quite similar. Therefore, for Random-Token Masking, the model pays equal attention to learning from these two kinds of words. However, the experimental results in Figure \ref{fig:pos} show that the language model dissipates its effort to model some function words, of which losses have been very low. Meanwhile, Random-Token Masking lets the model less likely learn those supposed-to-be-learn-more non-function words, which surely gives suboptimal pre-training consequences.

\section{Time-variant Masking Strategies}
\begin{figure*}[h]
	\centering
	\includegraphics[width=0.8\textwidth]{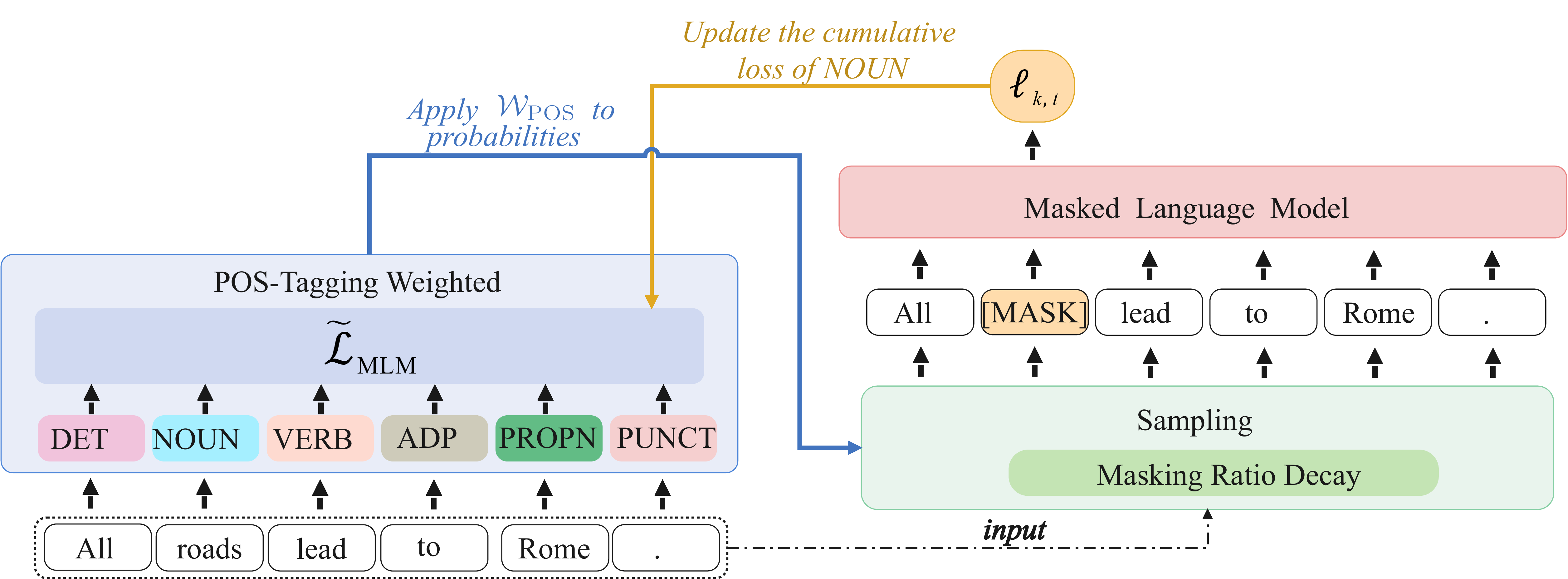}
	\caption{ Illustration of our time-variant masking.}
	\label{fig:framework}
\end{figure*}

In this section, we will present our exploration of time-variant masking on masking ratio and masked content inspired by our findings above. The overview of our time-variant masking is presented in Figure \ref{fig:framework}.
\subsection{Masking Ratio Decay (MRD)}
According to the observation in Section \ref{sec:mlm_ratio}, we design an optimized Masking Ratio Decay (MRD) schedule. At the beginning of pre-training, we use a high masking ratio and decay the masking ratio using certain strategies, which is similar to learning rate decay without warmup. Assuming that the model generally adopts a fixed masking ratio $p\%$ for training, we use a very high masking ratio (about $2p\%$) as the starting point and a very low masking ratio (nearly zero) as the ending point in MRD. 

\subsubsection{Implementation of Two Decay Methods}
\begin{figure}[h]
\centering
\includegraphics[width=0.4\textwidth]{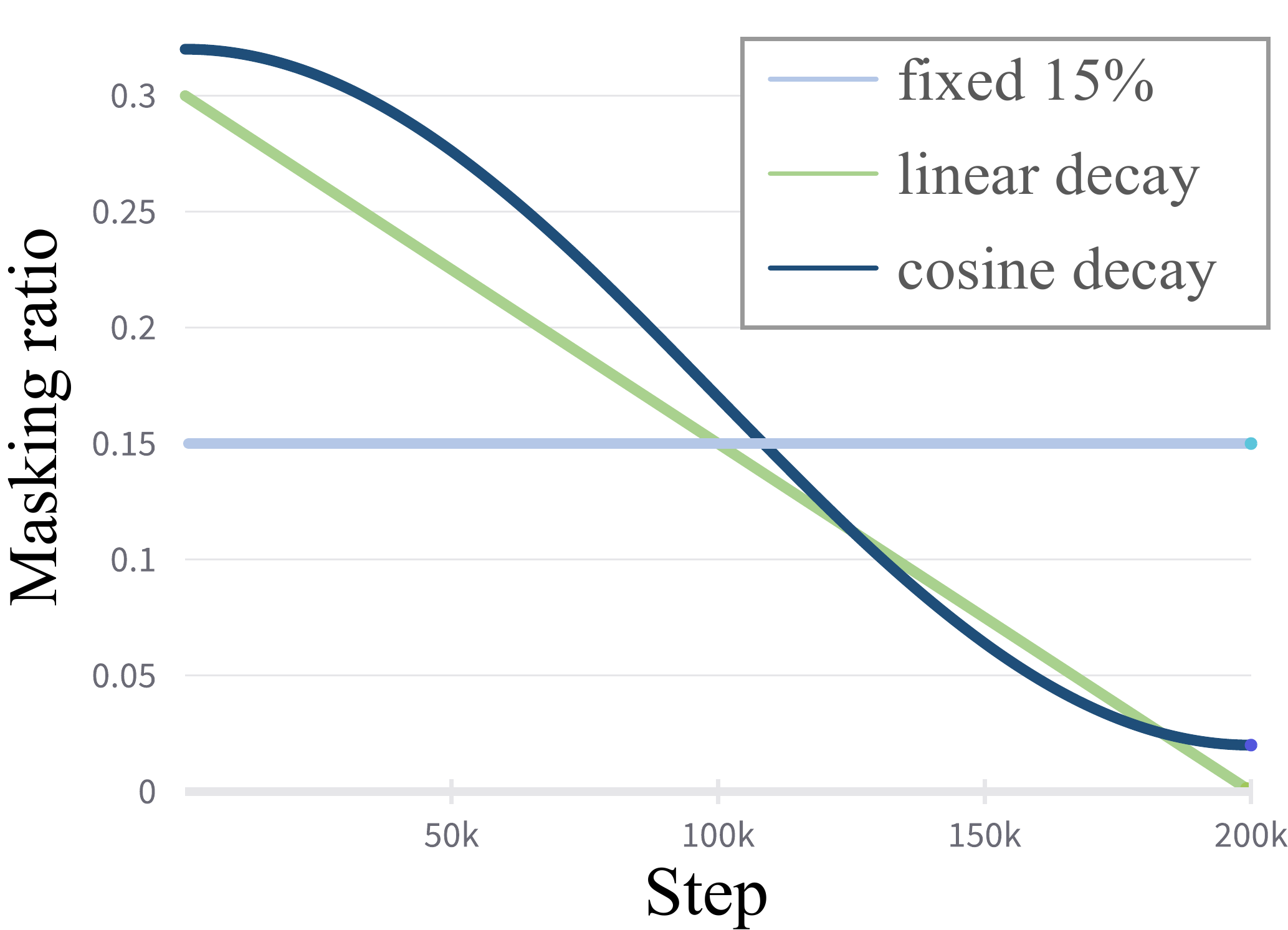}
\caption{Masking Ratio Decay schedules ($p=15$).}
\label{fig:mrd}
\end{figure}

We have tried two kinds of MRD to dynamically adjust the masking ratio, namely linear decay and cosine decay as follows:
\begin{align}
\label{eq:cos_lin}
\mathcal{M}_{linear}(t) & = (1 - \frac{t}{T}) \cdot 2p\% , \\
\mathcal{M}_{cosine}(t) & = (1 + cos(\frac{\pi}{T}t)) \cdot p\% + 0.02 ,
\end{align}
where $\mathcal{M}(t)$ is the current masking ratio at training step $t$ and $T$ is the total training step. Linear decay starts at $2p\%$ and decays to 0, while cosine decay starts at $2p\%+0.02$ and decays to 0.02, as shown in Figure \ref{fig:mrd}.
 
\subsubsection{Details of Design Intention}
We choose the starting point of $2p\%$ and ending point of 0 because the model using MRD can learn almost the same number of masked tokens as the baseline using a fixed masking ratio $p\%$ due to the central symmetry of linear and cosine functions for fair comparisons. The reason why we add 0.02 to the cosine decay is that the value of cosine function (masking ratio) is nearly 0 in the final 5\% steps, which means there are no masked tokens for model  to learn (and loss diminishes to 0). Thus we set a small number (0.02) to make the model keep training in the final stage.

\subsection{Analysis for How MRD Works}\label{sec:mrd_work}
MRD reminds us of the Simulated Annealing (SA) algorithm \citep{kirkpatrick1983optimization}, which is a greedy algorithm for optimization. In the SA algorithm, the degree of acceptance of suboptimal solutions depends on the annealing temperature $T$ according to the Metropolis algorithm \citep{metropolis1953equation}. That is, the higher the temperature parameter $T$ is, the larger the solution space allowed to be explored. Thus, the model can easily jump out the local minima if the $T$ is large. As the model converges and the annealing temperature decreases, the intolerance of suboptimal solution rises and a better local optimal solution can be found. In MRD, the magnitude of annealing temperature $T$ can be analogous to the masking ratio $p\%$. A high masking ratio means less information in the input sequence, allowing the model to explore more possibilities on coarse-grained task. On the other hand, a low masking rate allows the model to focus on finding a better solution close to global minima on the fine-grained task.

\subsection{POS-Tagging Weighted (PTW) Masking}
\label{sec:ptw_masking}
In this section, on the basis of the discussion of Section \ref{sec:pos}, we present the POS-Tagging Weighted (PTW) Masking, making the models have more chance to train on the difficult words according to the current training state.


Firstly, in the data pre-processing part, we use POS-tagging tools to label the words with corresponding part-of-speech. We then use WordPiece Tokenizer to perform tokenization and align tokens with their part-of-speech tags while ignoring the special tokens $[\mathrm{CLS}]$, $[\mathrm{SEP}]$, and $[\mathrm{PAD}]$.

Before training batch starts, we first apply PTW Masking to corrupt the sequences, while the masking ratio remains unchanged. When the model is trained at $t$ steps, according to Equation \ref{eq:pos}, we can obtain cumulative loss vector $\tilde{\mathcal{L}}_{\mathrm{MLM}}=\{\tilde{\ell}_{1}, \tilde{\ell}_{2}, \ldots, \tilde{\ell}_{k}, \ldots, \tilde{\ell}_{m}\},k \in \mathcal{C}$ for $m$ categories of words, and the cumulative loss vector $\tilde{\mathcal{L}}_{\mathrm{MLM}}$ is converted into the corresponding weight vector $\mathcal{W}_{\mathrm{POS}} = \{w_{1}, w_{2},\dots, w_{k}, \dots, w_{m}\},w_{k} \in (0,1)$ by the following equation:
\begin{equation}\label{eq:pos_wt}
    \mathcal{W}_{\mathrm{POS}} = %
    \mathrm{sigmoid}\Biggl(\frac{\tilde{\mathcal{L}}_{\mathrm{MLM}} - %
    \mathbb{E}(\tilde{\mathcal{L}}_{\mathrm{MLM}})} %
    {\sqrt{Var(\tilde{\mathcal{L}}_{\mathrm{MLM}})} \cdot \mu}\Biggr),
\end{equation}
where $\mu$ is a coefficient to adjust the input for sigmoid function. We apply this weight vector $\mathcal{W}_{\mathrm{POS}}$ to the masking probabilities. {Equation \ref{eq:pos_wt}}\allowbreak is based on Equation \ref{eq:pos}, where the process of smoothing gives the weights changing relatively stably for masking. We do not use bias correction in Equation \ref{eq:pos}, which enables the cumulative losses for each kind of words to grow from zero. That is, in the very beginning, the $\mathcal{W}_{\mathrm{POS}}$ is initialized with the same value for each type of words and weights the probabilities for masking equally.

Specifically, the masking probability of each word is weighted by its corresponding part-of-speech, so that words with higher losses are more likely to be masked. We show that non-function words tend to have much higher losses than function words, so the language model learns to model non-function words most of the time, but fewer function words and punctuation, as shown in Figure \ref{fig:weights}. In special case, PTW Masking is similar to Named Entities Masking \citep{sun2019ernie} if only proper nouns have a weight of 1 and the others are 0.


\section{Experiment Setup}
\label{sec:exp}
\subsection{Datasets and Setup}
For full details of pre-training and finetuning, please see the Appendix \ref{appendix:details}.
\subsubsection{Pre-training}
\label{sec:pre-train}
For pre-training, we use the BERT-base and  BERT-large model as the representatives of MLMs for training. Specifically, we train BERT-base for 200k steps from scratch and continue training BERT-large models for 100k steps initialized by pre-trained weights \footnote{\url{https://huggingface.co/bert-large-uncased}}. In practice, we explore masking ratio with MRD and masked content with PTW Masking seperately to avoid mutual influence. For the dataset, we train BERT on English Wikipedia using WordPiece Tokenizer for tokenization. 

\subsubsection{Finetuning}
We finetune our models on GLUE \citep{wang-etal-2018-glue} and SQuAD v1.1 \citep{Rajpurkar2016SQuAD} to evaluate the performance of downstream tasks. To further explore what extra skills models using PTW Masking have learnt, we evaluate BERT-large on CoNLL-2003 \citep{tjong-kim-sang-de-meulder-2003-introduction}, which is a widely used NER benchmark, to test the information extraction ability of the models. We train all tasks 5 times respectively and report the average scores. For more detailed introduction of downstream tasks, please see Appendix \ref{appendix:benchmark}.

\begin{table*}[ht]
\centering
\setlength{\tabcolsep}{5.5pt}
\caption{Evaluating the performance of models using MRD but trained with two different total training steps. For GLUE, we report STS using Spearman correlation and CoLA using Matthew’s correlation, and other tasks using accuracy. And we report SQuAD using F1.}
\label{tab:mrd_result}
 \small
{
\begin{tabular}{lcccccccccc}
\toprule
{\textbf{Model}} & \textbf{CoLA} & \textbf{SST} & \textbf{MRPC} & \textbf{STS} & \textbf{QQP} & \textbf{MNLI} & \textbf{QNLI} & \textbf{RTE} & \textbf{GLUE Avg}  & \textbf{SQuAD}\\
\midrule
Masking Ratio Decay & & & & \\
\midrule
\emph{Trained for 200k steps} \\
\quad Random-Token 15\%            & 43.8 & 90.4 & 85.6 & 85.9 & 90.0 & \textbf{81.5} & 88.7 & 62.2 & 78.5 & 87.1 \\
\quad Linear Decay                 & 44.5 & \textbf{90.5} & \textbf{86.0} & 85.4 & 90.0 & 81.0 & 89.6 & \textbf{63.2} & 78.8($\uparrow$0.3) & 87.7($\uparrow$0.6)\\
\quad Cosine Decay                 & \textbf{44.9} & 90.3 & 85.8 & \textbf{86.1} & \textbf{90.2} & 81.2 & \textbf{89.7} & 62.6 & \textbf{78.9}($\uparrow$0.4) & \textbf{87.7}($\uparrow$0.6)\\
\emph{Trained for 1M steps}   \\
\quad Random-Token 15\%            & 54.8 & 91.5 & 86.2 & 86.8 & 90.3 & 83.7 & 90.8 & 64.7 & 81.1 & 90.2\\
\quad Linear Decay                 & 55.2 & 92.3 & 86.8 & 87.7 & 90.4 & 84.4 & 91.2 & \textbf{70.4} & 82.3($\uparrow$1.2)  & 90.6($\uparrow$0.4)\\
\quad Cosine Decay                 & \textbf{57.2} & \textbf{92.3} & \textbf{88.2} & \textbf{87.8} & \textbf{90.5} & \textbf{84.7} & \textbf{92.3} & 69.5 & \textbf{82.8}($\uparrow$1.7) & \textbf{90.9}($\uparrow$0.7)\\
\bottomrule
\end{tabular}
}
\end{table*}

\subsection{Implementation Details}
\label{sec:method}
For the implementation of MRD, we train the model using Random-Token Masking with a fixed masking ratio of 15\% as the baseline. Then we apply our MRD to Random-Token Masking with an average masking ratio of 15\% ($p=15$) using linear and cosine decay, respectively. 

\begin{figure}[ht]
	\centering
	\includegraphics[width=0.35\textwidth]{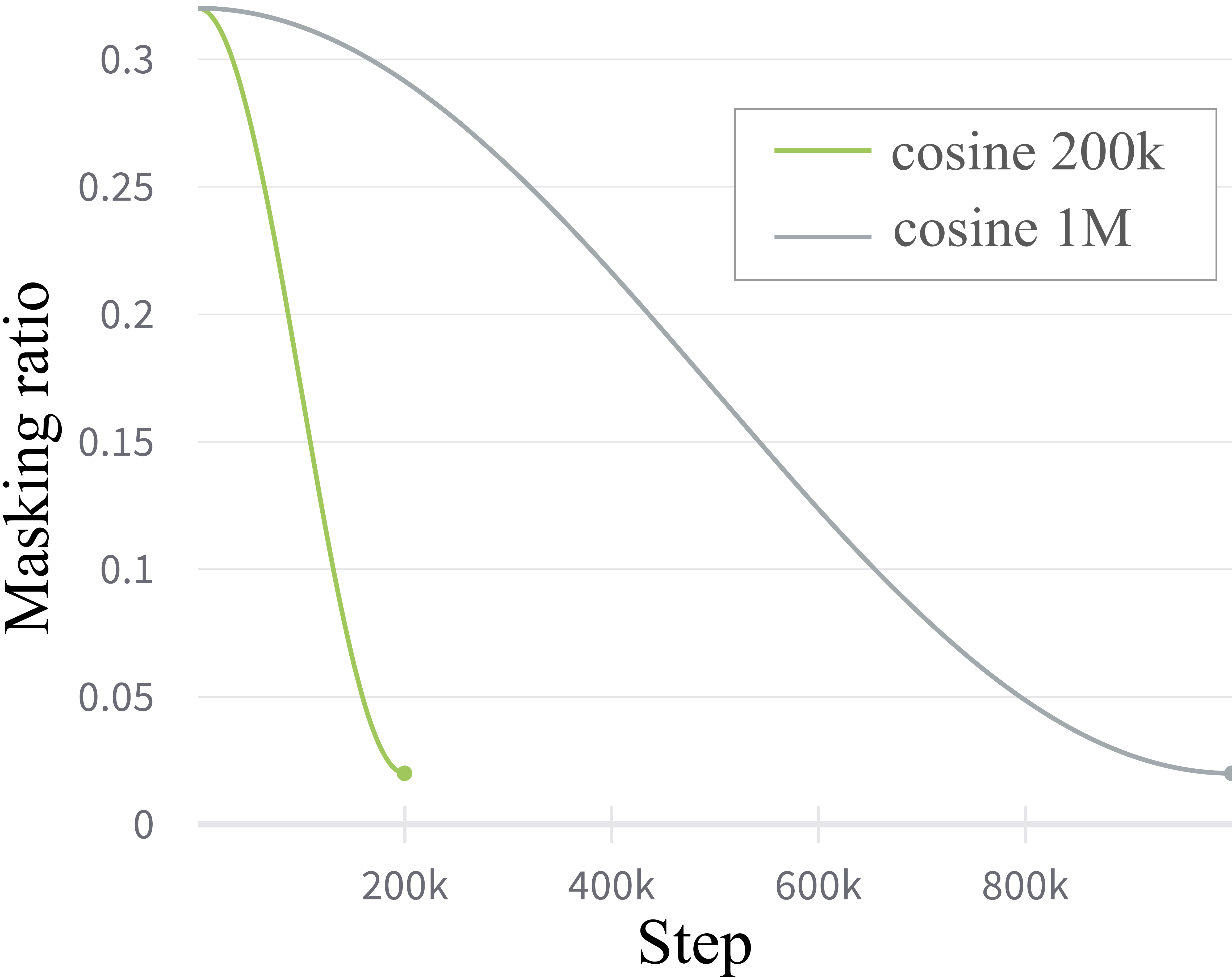}
	\caption{Different total training steps cause an absolute difference in masking ratio in the early stage of pre-training though same MRD strategies are applied.}
	\label{fig:200k_1m}
\end{figure}

Because there is the parameter $T$ in Equation \ref{eq:cos_lin}, the number of training batches learned by the models under a certain masking ratio will differ if total training step $T$ is different. As shown in Figure \ref{fig:200k_1m}, compared to the model with 200k steps, the models with 1M steps are trained for large masking ratio for longer time in early stage. This question will not be raised if we use the fixed masking ratio as we usually do. But in MRD, though both masking ratios decay in relatively the same way, the absolute difference of masking ratio in early stage may affect the performance of downstream task. Thus, we additionally train the models for 1M steps to explore if training on large masking ratio longer time or decaying faster at early stage is more beneficial to pre-training.

For the implementation of PTW Masking, we also use Random-Token Masking with a fixed masking ratio of 15\% as baseline. We train the model with PTW Masking with the same fixed ratio of 15\% for equal comparisons.


\begin{table*}[!ht]
\centering
\setlength{\tabcolsep}{4.3pt}
\caption{Evaluating the performance of models using PTW Masking. We report CoNLL using F1.}
\label{tab:ptw_result}
 \small
{
\begin{tabular}{lccccccccccc}
\toprule
{\textbf{Model}} & \textbf{CoLA} & \textbf{SST} & \textbf{MRPC} & \textbf{STS} & \textbf{QQP} & \textbf{MNLI} & \textbf{QNLI} & \textbf{RTE} & \textbf{GLUE Avg}  & \textbf{SQuAD} & \textbf{CoNLL}\\
\midrule
PTW Masking & & & & \\
\midrule
\multicolumn{2}{l}{$\mathrm{\textbf{BERT}_{base}}$ \emph{from scratch}} \\
\quad Random-Token            
& 43.8 & \textbf{90.4} & \textbf{85.6} & 85.9 & 90.0 & 81.5 & 88.7 & 62.2 & 78.5 & 87.1 & - \\
\quad PTW      
& \textbf{45.3} & 90.0 & 85.4 & \textbf{86.8} & \textbf{90.1} & \textbf{82.0} & \textbf{91.1} & \textbf{62.5} & \textbf{79.2}($\uparrow$0.7) & \textbf{88.1}($\uparrow$1.0) & - \\
\multicolumn{4}{l}{$\mathrm{\textbf{BERT}_{large}}$ \emph{from continue-training}}  \\
\quad Random-Token          
& 61.8 & 92.8 & 86.5 & 89.3 & 91.3 & 86.3 & 92.4 & 67.2 & 83.4 & 91.3 & 94.9 \\
\quad PTW  
& \textbf{62.5} & \textbf{93.3} & \textbf{86.8} & \textbf{90.0} & \textbf{91.3} & \textbf{86.6} & \textbf{92.5} & \textbf{68.2} & \textbf{83.9}($\uparrow$0.5) & \textbf{91.6}($\uparrow$0.3) & \textbf{95.4}($\uparrow$0.5)\\

\bottomrule
\end{tabular}
}
\end{table*}


\section{Results}

\subsection{Results of MRD}
The experimental results show that MRD greatly improves downstream task performance and pre-training efficiency (and more discussion on MRD in Appendix \ref{appdix:add_mrd}).

\subsubsection{Decaying Masking Ratio vs. Fixed Ratio}
\begin{figure}[t]
	\centering
	\includegraphics[width=0.4\textwidth]{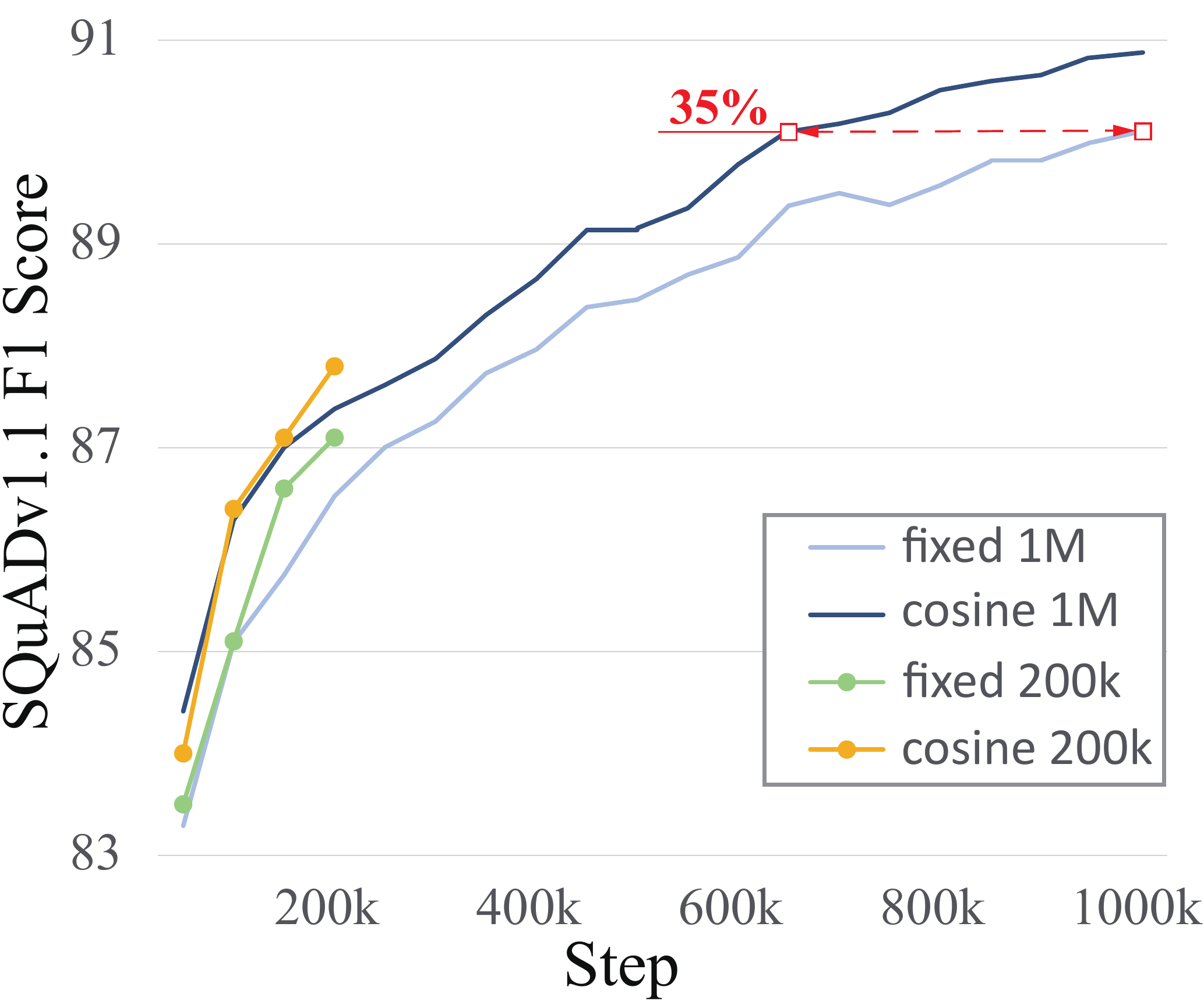}
	\caption{Comparisons between fixed ratio and cosine decay strategy on SQuAD performance during pre-training. We evaluate the saved checkpoints for every 50k steps on SQuAD v1.1 dev set following the same experimental setup.}
	\label{fig:squad_mrd}
\end{figure}

In Figure \ref{fig:squad_mrd}, we show the SQuAD performance for every 50k checkpoint during pre-training. We observe that the large masking ratio gives a better downstream performance at the start and the decaying mechanism continues to boost the downstream performance, which takes the advantage of high masking ratio and low masking ratio discussed in Section \ref{sec:analysis}. The model using cosine decay at 650k steps has obtained a competitive SQuAD v1.1 F1 score to the baseline at 1M steps, thus reducing the training time by 35\%. 

\subsubsection{Influence of Masking Ratio at Early Stage}
We further explore the absolute difference (mentioned in Section \ref{sec:method}) with different training steps in masking ratio using the same MRD strategies. As shown in Table \ref{tab:mrd_result}, for GLUE tasks, MRD training for 1M steps has more obvious advantages than 200k steps. The model trained with 1M steps performs well above baseline on all subtasks using MRD, with an average increase of 1+ on GLUE, which has a larger increment compared to 200k steps. The comparison shows that models benefit from training for a longer time with a large masking ratio from the start, especially on GLUE. Because the subtasks in GLUE are mainly sequence-level, which focus on global semantics. For a higher masking ratio, the model tries to train on a coarse-grained task, inferring global semantics from fewer words, which is more suitable for GLUE. Therefore, in the training of 200k steps, the masking ratio decays too fast, resulting in insufficient training on coarse-grained tasks. In contrast, the model with 1M steps can maintain the training at a high masking ratio for a longer time and thus perform better in the sequence-level tasks of GLUE. 

\subsubsection{Comparisons Between Different MRD}
Compared with linear decay, cosine decay has better downstream task performance in most subtasks in GLUE. The difference between these two is that cosine decay keeps a higher masking ratio in the early stage and decays more quickly, which is consistent with the analysis mentioned above. To move forward, it is necessary to maintain a high masking ratio in the early stage of pre-training. According to Section \ref{sec:mrd_work}'s empirical analysis, the model can explore a larger global solution space by using a higher masking ratio in the early training period so as to better converge to the optimal global minima when the masking ratio decreases later.

\subsection{Results of PTW Masking}
Results in Table \ref{tab:ptw_result} show that PTW Masking has significantly improved the performance of various downstream tasks. 

\subsubsection{What Skills Models Have Learnt if Trained with Mostly Non-function Words with PTW Masking?}
Because models trained with PTW Masking pay more attention to important words like non-function words, we can see that models are especially good at information extraction tasks, e.g., extractive QA and NER. The gains in SQuAD and CoNLL in Table \ref{tab:ptw_result} have shown models are sensitive to words with more semantic information, which is consistent with the design of the pre-training goal. 

Besides that, PTW Masking also does well in several NLU tasks in GLUE. PTW Masking is a time-variant strategy, which aggregates the advantages of both Random-Token Masking and Named Entities Masking. From the start of pre-training, models can learn from all the words  equiprobably like Random-Token Masking. And at the later stage, models memorize more knowledge by masking important words instead of wasting time on predicting meanless words. Thus models can have better NLU ability with memorization of more knowledge under the condition of training with the same number of tokens.


\section{Related Work}

The pre-processing of the MLM is to replace a subset of the tokens in the input with $\mathrm{[MASK]}$ tokens, which has two considerations to optimize: how many tokens to mask (masking ratio) and what tokens to mask (masked content). 

\textbf{$\bullet$ Masking Ratio}. 
Masking ratio is a very important hyperparameter that affects the pre-training of MLM, which is relatively seldom studied. In BERT, the masking ratio of 15\% is the most commonly used value and is also applied in other MLMs. The generator of ELECTRA \citep{clark2019electra} is a MLM, using 15\% for base-sized models and 25\% for large-sized models. However, considering the cooperation with the discriminator, it is difficult to judge the effect of 25\% on MLM. In a recent study, \cite{wettig2022should} suggests that a masking ratio of 40\% performs better than 15\% in downstream tasks of RoBERTa-large \citep{liu2019roberta} model. T5 \citep{raffel2020exploring} uses an MLM-style pre-training method and also experiments on the influence of different masking ratios. They find that the masking ratio has a limited effect on the model's performance except for 50\% and use 15\% as the final choice. To our best knowledge, most studies on masking ratio compare the performances of downstream tasks at the end of pre-training \citep{raffel2020exploring}, but few studies pay attention to the dynamic influence of masking ratio during pre-training, which is very interesting. We record the changes in the performance of downstream tasks under different masking ratios and therefore propose the MRD according to the empirical law we observe. Instead of using a fixed masking ratio, we dynamically decay the ratio and find that the performance of MLM can be greatly improved.

\textbf{$\bullet$ Masked Content}. 
Previous studies have explored strategies for masked content to further improve the Random-Token Masking, though nearly all of them focus on how to select coherent enough masked units.
\cite{devlin-etal-2019-bert} proposes Whole-Word Masking, which forces the model to predict complete words instead of WordPiece tokens. Furthermore, SpanBERT \citep{joshi-etal-2020-spanbert}, $n$-gram Masking \citep{levine2021pmimasking,li-zhao-2021-pre} and LIMIT-BERT \citep{zhou2020limit} take into account the continuous mask of multiple word combinations, making model predict tokens using the context with long dependencies. 
ERNIE \citep{sun2019ernie} improves pre-training performance by especially masking named entities. Different from all existing MLM improvements, our PTW Masking lets different types of words correspondingly receive the matched learning intensity, which pioneers a new technical line for the concerned MLM.

\section{Conclusion}
\label{sec:conclude}
Masked language model pre-training can be generally defined by two main factors, masking ratio and masked contents. The Random-Token Masking scheme adopted by existing studies treats all words equally and maintains a fixed ratio throughout the entire pre-training, which has been shown suboptimal in our analysis. 
To better unleash the strength of MLM, we explore two kinds of time-variant masking strategies, namely, Masking Ratio Decay (MRD) and POS-Tagging Weighted (PTW) Masking. Experimental results verify our hypothesis that MLM benefits from time-variant setting both in masking ratio and masked content according to dynamic training states.
Our further analysis show that these two time-variant masking schedules greatly improve pre-training efficiency and the performance of downstream tasks. 




\section*{Limitations}
One limitation of this work is that both PTW Masking and MRD are conducted only on BERT due to limited resources, and MLMs with other structures may have different reactions to the time-variant masking with different contents and ratios. Another limitation is that although we propose MRD for the first time, the strategy of time-variant masking ratio is hard to design like learning rate decay. In fact, other decay methods and choices of starting and ending point are various, where better strategies may exist and further work can be done.

\bibliography{custom}
\bibliographystyle{acl_natbib}

\appendix
\section{Additional Investigation on MRD}
\label{appdix:add_mrd}
\subsection{Magnitude of Masking Ratio in MRD}
When using MRD, we explore the influence of the much higher masking ratios, which affect the downstream performance of the model. Previous studies \citep{raffel2020exploring, wettig2022should} have shown that a much higher fixed masking ratio ($\gg 40\%$) will cause significant degradation in the model performance because the model can only infer from a small amount of known information resulting in quickly converging to local minima. In MRD, we show that the design of the decaying mechanism can mitigate the impact of the high masking ratio. For the BERT-base model, starting from a high ratio ($30\%$) and a much higher ratio ($55\%$), both can outperform the baseline with a similar margin. We show that higher masking ratios in early pre-training stage help downstream performance, and MRD prevents the high masking ratios from destroying pre-training in later stage.

\subsection{MRD Interacts with Learning Rate}
Moreover, we show a subtle relationship between MRD and learning rate decay. When the masking ratio is low, using a relatively high learning rate will cause a huge decline in model performance. Therefore, in MRD, the masking ratio and the learning rate both adopt the same type of decay strategy except that the learning rate has an additional warmup stage. For example, cosine masking ratio decay uses cosine learning rate decay.

\subsection{Other Simple Schedules in MRD}
\begin{figure}[h]
	\centering
{
	\includegraphics[width=0.4\textwidth]{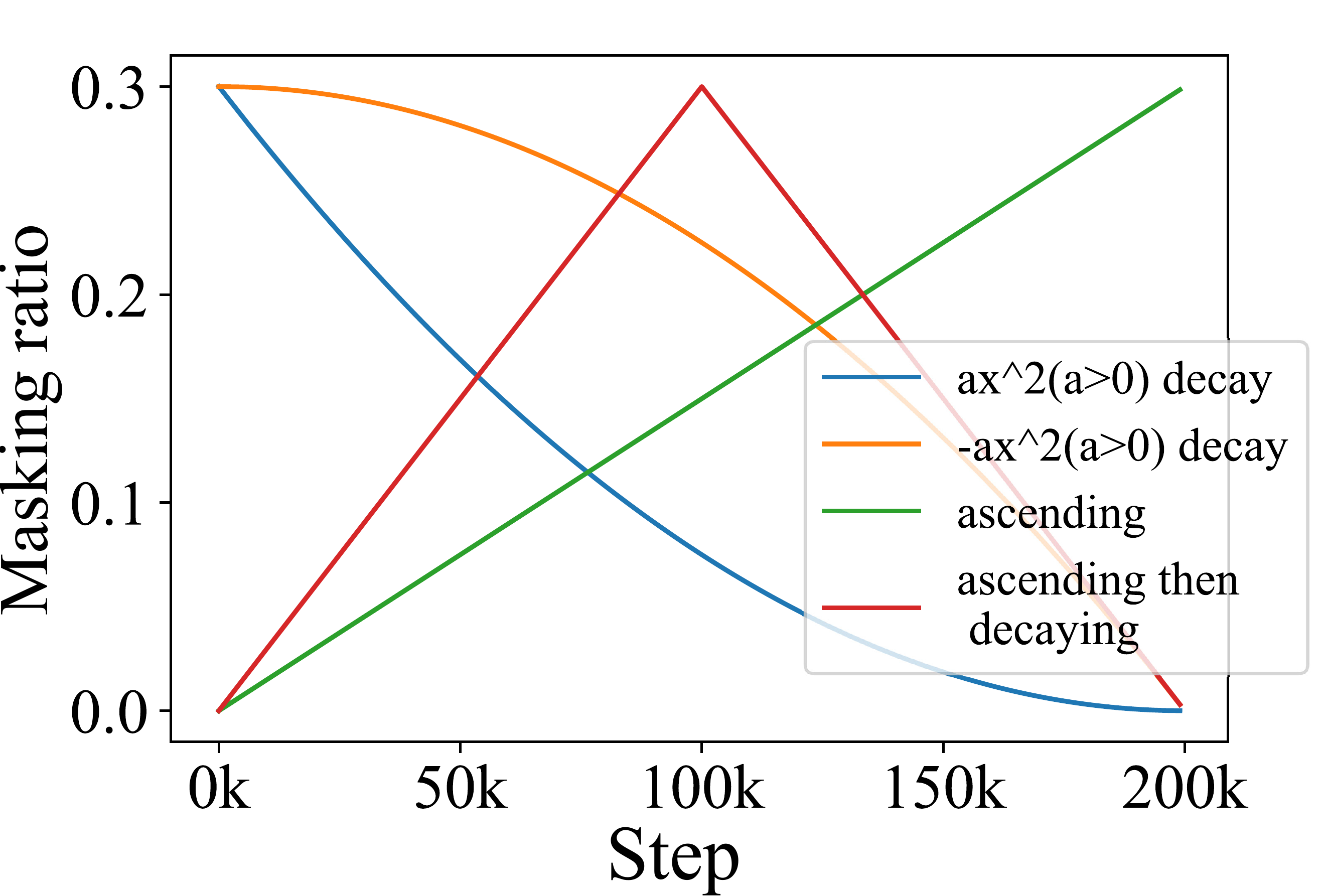}
	\caption{Other simple schedules of adjusting the masking ratios. }
	\label{fig:simple_ratio}
}
\end{figure}

Based on the same experiment setup, we train the models with other simple schedules (shown in Figure \ref{fig:simple_ratio}) for 200k steps using the linear learning rate decay and finetune them on the SQuAD v1.1. The results on SQuAD v1.1 dev set are presented in Table \ref{tab:simple_squad}. We find that cosine is the best compared with those alternatives.


\begin{table}[h]
\centering
{
\setlength{\tabcolsep}{2.8pt}
\begin{tabular}{lcccc}
\toprule
\multirow{2}{*}{\textbf{Model}}
& \multicolumn{2}{c}{\textbf{SQuAD v1.1}}
\\
& EM  & F1 
\\
\midrule
Masking Ratio Decay & &  \\
\midrule
\quad Random 15\%               & 79.4  & 87.1  \\
\quad $-ax^2 (a>0)$ Decay        & 79.8  & 87.6 \\
\quad $ax^2 (a>0)$ Decay       & 79.3  & 87.2  \\
\quad Ascending                 & 79.6  & 87.4 \\
\quad Ascending then Decaying   & 79.4  & 86.9 \\
\quad Linear Decay              & 79.8  & 87.7 \\
\quad Cosine Decay              & \textbf{79.9}  & \textbf{87.7} \\
\bottomrule
\end{tabular}
}
	\caption{Results of other simple schedules on the SQuAD v1.1 dev set.
	}
	\label{tab:simple_squad}

\end{table}

\section{Finetuning Benchmarks}
\label{appendix:benchmark}
\paragraph{GLUE}
The General Language Understanding Evaluation (GLUE) benchmark is a collection of 9 various tasks for sequence-level classification for evaluating natural language understanding systems. 

\paragraph{SQuAD}
The Stanford Question Answering Dataset (SQuAD) is a commonly used benchmark for question answering. The task is to predict the text span of an answer from a given passage-question pair. 
\paragraph{CoNLL}
The CoNLL-2003 concerns language-independent named entity recognition. It concentrates on four types of named entities: persons, locations, organizations and names of miscellaneous entities that do not belong to the previous three groups.

\section{Pre-training and Finetuning Details}
\label{appendix:details}
\subsection{Pre-training Details}
We only use the MLM task as the training objective and discard the Next Sentence Prediction task, as it has been shown to be redundant in previous studies \citep{liu2019roberta, joshi-etal-2020-spanbert}. 

\begin{table}[h]
\centering
{
\setlength{\tabcolsep}{10pt}
\begin{tabular}{lcc}
\toprule
\textbf{Hyperparameter} & \textbf{Base} & \textbf{Large} \\
\midrule
Learning Rate & 2e-4 & 5e-5 \\
Warmup Steps & 10000 & 10000\\
Weight Decay & 0.01 & 0.01 \\
Batch size & 256 & 256 \\
Sequence Length & 512 & 512 \\ 
Gradient Clipping & 1.0 & 1.0 \\
\bottomrule
\end{tabular}
}
\caption{Pre-training hyperparameters for BERT models.}

\label{tab:pretrain_hyp}
\end{table}

\subsection{Finetuning Details}
Following the common finetuning practice, we do not use any additional training strategies. We train all tasks 5 times respectively and report the average scores. For GLUE, We use 8 widely used tasks in GLUE.\footnote{
 For a fair comparison, we exclude the WNLI following the previous work \citep{devlin-etal-2019-bert}.}
\begin{table}[h]
\centering
{
\setlength{\tabcolsep}{5pt}
\begin{tabular}{lcc}
\toprule
\textbf{Hyperparameter} & \textbf{Base} & \textbf{Large} \\
\midrule
GLUE \\
\midrule
Learning Rate & \{5e-5, 1e-4\} & \{1e-5, 2e-5\} \\
Batch Size & 32 & 32 \\
Weight Decay & 0 & 0 \\
Training Epochs* & 3 & 3 \\
\midrule
SQuAD v1.1 \\
\midrule
Learning Rate & \{5e-5, 1e-4\} & \{2e-5, 5e-5\} \\
Batch Size & 128 & 32 \\
Weight Decay & 0 & 0 \\
Training Epochs & 3 & 3 \\
\midrule
CoNLL-2003 \\
\midrule
Learning Rate & - & \{2e-5, 5e-5\} \\
Batch Size & - & 32 \\
Weight Decay & - & 0 \\
Training Epochs & - & 3 \\
\bottomrule
\end{tabular}
}
\caption{Finetuning hyperparameters for BERT models.\\
*We finetune our models in RTE and STS-B for 10 epochs and other subtasks for 3 epochs.}
\label{tab:finetune_hyp}
\end{table}

\end{document}